\documentclass{article}

\usepackage{PRIMEarxiv}

\usepackage[utf8]{inputenc} 
\usepackage[T1]{fontenc}    
\usepackage{hyperref}       
\usepackage{url}            
\usepackage{booktabs}       
\usepackage{amsfonts}       
\usepackage{nicefrac}       
\usepackage{microtype}      
\usepackage{lipsum}
\usepackage{fancyhdr}       
\usepackage{graphicx}       
\graphicspath{{media/}}     

\title{A Deep Learning Approach for Masking Fetal Gender in Ultrasound Images
}

\author{
    Amit Borundiya \\
    \texttt{amit@launchpad.ai} \And
    Arshak Navruzyan \\
    \texttt{arshak@fellowship.ai} \And 
    Dennis Igoschev \\ 
    \texttt{denisigoshev@gmail.com} 
    \And 
    Feras C. Oughali \\
    \texttt{oughali.feras@gmail.com} \And 
    Hemanth Pasupuleti \\
    \texttt{satyasaihemanth.p@gmail.com} \And 
    Mike Fuller \\
    \texttt{mikefullergm@gmail.com} \And 
    Vinay Kanigicherla \\
    \texttt{vinaykanigicherla@berkeley.edu} 
    \And 
    T S Aniruddha Kashyap \\
    \texttt{aniruddhaani93@gmail.com} \And 
    Rishabh Chaurasia \\
   \texttt{chaurasia.rishabh7@gmail.com} 
    \And
    Dr. Sonali Vinod Jain \\
    \texttt{drsonalijain174@gmail.com}
  \\ 
  \\

 \centerline{Fellowship.AI}

}

\begin{document}
\maketitle

\begin{abstract}
Ultrasound (US) imaging is highly effective with regards to both cost and versatility in real-time diagnosis; however, determination of fetal gender by US scan in the early stages of pregnancy is also a cause of sex-selective abortion. This work proposes a deep learning object detection approach to accurately mask fetal gender in US images in order to increase the accessibility of the technology. We demonstrate how the YOLOv5L architecture exhibits superior performance relative to other object detection models on this task. Our model achieves 45.8\% $AP_{[0.5:0.95]}$, 92\% F1-score and 0.006 False Positive Per Image rate on our test set. Furthermore, we introduce a bounding box delay rule based on frame-to-frame structural similarity to reduce the false negative rate by 85\%, further improving masking reliability.
\end{abstract}


\section{Introduction}
\label{sec:introduction}
In the medical domain, there are a multitude of options with regards to tomography diagnostic technologies such as X-Rays, Magnetic Resonance Imaging (MRI), Ultrasound (US), and Computed Tomography (CT). Among all the diagnostic imaging options, Ultrasound imaging is highly effective in terms of cost and its versatility in medical diagnosis on various parts of the body in real-time. While US imaging is used worldwide, it has a tremendous impact in developing countries as its portability and cost-effectiveness make it one of the only widely accessible imaging technologies.

However, US imaging faces some challenges in many developing countries where the technology is perceived to be responsible for infant foeticide as it can be used to determine the gender of the fetus early in a pregnancy. To demonstrate the significance of sex-selective abortion, a study \cite{sex:selective:abortion} was conducted to provide information on sex ratio at birth (SRB) reference levels and SRB imbalance. The study identified 12 countries with strong statistical evidence of SRB imbalance. To quantify the effect of SRB imbalance due to sex-selective abortion, the annual number of missing female births (AMFB) and the cumulative number of missing female births (CMFB) were calculated over time. Missing female birth estimates for the 12 countries from 1970 to 2017 is 23.1 million. The majority of CMFB are concentrated in China with 11.9 million and in India with 10.6 million. The CMFB between 1970 and 2017 in China and India made up 51.40\% and 45.94\%, respectively, of the total CMFB.

In order to combat sex-selective abortion, many of these countries have stringent laws regulating the use of US machines. Though put in place for good reason, these laws hinder the aforementioned accessibility that makes US imaging such a powerful and versatile technology.

The main contribution of this study is the development of a novel deep learning method that can mask the fetal genital region in US imaging video in real-time and thus prevent the detection of fetal gender in the early days of pregnancy by anyone except an authorized medical practitioner. Solving this challenge will enable further adoption of US technology in developing countries while simultaneously preventing the occurrence of sex-selective abortions.

The paper is organized as follows. Section \ref{sec:related_work} presents presents a literature review of related work. Section \ref{sec:dataset} provides details about the dataset and its creation. Section \ref{sec:exp} reports the results of experimental work and provides a performance comparison of various DNN architectures. Section \ref{sec:opt} investigates parameter optimizations to reduce the occurrence of false positive and false negative detections. Section \ref{sec:Sono} demonstrates the performance of our model's masking as reported by an expert sonographer. 
Section \ref{sec:further} details potential further work. Lastly, Section \ref{sec:con} concludes this study.

\section{Related Work}
\label{sec:related_work}

The clinical value of identification of fetal gender by ultrasound lies in deciding whether to carry out prenatal invasive testing in pregnancies at the risk of sex-linked genetic abnormalities. A study \cite{gender_id_1} was carried out to assess the accuracy of fetal gender determination at 11–14 weeks of gestation. Results showed that the accuracy of gender determination increased with gestation from 70.3\% at 11 weeks to 98.7\% at 12 weeks and 100\% at 13 weeks. Male fetuses were wrongly assigned as female in 56\% of cases at 11 weeks, 3\% at 12 weeks and 0\% at 13 weeks. In contrast, only 5\% of the female fetuses at 11 weeks were incorrectly assigned as male, and this false-positive rate was 0\% at 12 and 13 weeks.

Another similar study \cite{gender_id_2} was performed on 496 singleton pregnancies in the first trimester. It was possible to determine fetal gender for 441 out of 496 patients. Of the 55 cases where no identification of gender was possible, 39 were in the 11-week gestational age group, representing 40.6\% of this category. The attending obstetrician was a certified sonographer for first-trimester screening registered with the Fetal Medicine Foundation of England. Reported results show the success rate for correctly identifying fetal gender (where identification was possible) increased with gestational age, from 71.9\% at 11 weeks, 92\% at 12 weeks, and 98.3\% at 13 weeks.

Lakra et al. \cite{Lakra} proposed a deep learning-based method to identify in real-time frames of B-scans that contain gender revealing features and block those frames out from the display. A residual network with 23 convolutional layers was used. The accuracy of the classifier was found to be 83.56\% on average, with a 69.57\% F1-score using 5-fold cross-validation. Ultrasound B-mode fetal scan data were obtained from 50 women 12 to 20 weeks into their pregnancy. This time frame was selected as the gender of the fetus becomes identifiable from ultrasound scans during this period, and abortion in India is allowed until 20 weeks of pregnancy.

\section{Dataset}
\label{sec:dataset}

It is understood that there is no publicly available dataset of ultrasound images pertaining to fetal gender determination. Therefore, images of both \textit{transverse} and \textit{mid-sagittal} planes containing the fetal genital tubercle were collected from public pregnancy forums to compile the labeled data. In order to simulate a typical fetal US scan in which both gender identification and non-gender identification frames are visible, negative samples were taken from a recently published dataset of maternal-fetal ultrasound images \cite{nature}. The maternal-fetal dataset contains images from the most widely used fetal anatomical planes (Abdomen, Brain, Femur and Thorax), the mother’s cervix, and a general category to include any other less common image or plane. Images were selected based on patient ID for both validation and test sets to ensure there is no data leakage. Images from the classes "Femur" and "Other" of the maternal-fetal dataset were omitted as those classes contain some images with transverse and mid-sagittal features or irrelevant frames such as electrocardiogram plots. Negative samples in train, validation and test sets comprise 72\% of total images \cite{Lakra}. Furthermore, they have been selected using stratified sampling of the plane classes and sub-classes so as to maintain the class distribution of the original maternal-fetal dataset. The training set contains 70\% of both labeled images and negative samples, while validation and test sets both contain 15\% each of the labeled images and negative samples. Table~\ref{table:dataset} shows the composition of the dataset.

\begin{table}
\centering
\caption{Dataset Composition \strut}
\begin{tabular*}{0.6\textwidth}{cccccc}
\toprule
\multicolumn{1}{c}{Dataset} & \multicolumn{1}{c}{Total} & \multicolumn{3}{c}{Labeled} &  \multicolumn{1}{c}{Negative}\\
\cmidrule(lr){3-5}
& & Total & Transverse & Mid-Sagittal & \\
\midrule
Training & 2534 & 707 & 355 & 352 & 1827\\
Validation & 543 & 152 & 77 & 75 & 391\\ 
Test & 543 & 152 & 76 & 76 & 391\\ 
\bottomrule
\end{tabular*}
\label{table:dataset}
\end{table}

\subsection{Labeled Data Acquisition}

Images of \textit{transverse} and \textit{mid-sagittal} planes containing the fetal genital tubercle section were manually collected from public pregnancy forums \footnote{www.whattoexpect.com and www.ingender.com}. Only high-quality images were selected. Images that were small in size, blurry, or with artifacts such as a reflection of ambient light were discarded. A rigorous exercise to ensure no duplicate images was strictly followed.

\subsection{Data Preprocessing}

The collected images went through a pipeline of preprocessing steps to ensure that the dataset was of high quality. Images that had background noise near the borders were cropped appropriately. Images that had excessive text or inpainted shapes, such as arrows near the genital region, were discarded. In some cases, images with minor inpainted text or shapes were treated using a segmentation mask followed by applying OpenCV's \textit{inpaint} function. Due to imperfections in the segmentation masks, a top-hat morphological operation with a $7\times7$ rectangular structuring element was used, similar to~\cite{Efford2000DigitalIP} and the edges were recovered by dilating the image with a $3\times3$ rectangular structuring element. The thresholded text segmentation mask was obtained using OTSU thresholding~\cite{4310076} and Hysteresis thresholding~\cite{4767851}. 

\subsection{Data Annotation}

In this study, two classes of gender identification planes were considered: the \textit{transverse} plane and the \textit{mid-saggital} plane. An effort has been made to ensure that the dataset is balanced across considered classes. All bounding boxes were squares such that no obvious differences between the bounding box shapes are apparent between the sexes.

Regarding the annotation of the transverse plane: 
\begin{itemize}
  \item For females, the edge of the genitalia should sit in the center of the bounding box. For males, the edge of the penis should sit at the relative edge of the bounding box.
  \item The inner edges of either leg should be present towards the boundary of the bounding box.
  \item The genital region up to the perineum should be captured. 
\end{itemize}

Regarding the annotation of the mid-sagittal plane:
\begin{itemize}
  \item The entirety of the typically brighter tubercle length should be captured up to the edge of the bounding box.
  \item The bounding boxes should include part of the fetal body.
\end{itemize}

\begin{figure}[!t]
\centerline{\includegraphics[scale=0.3]{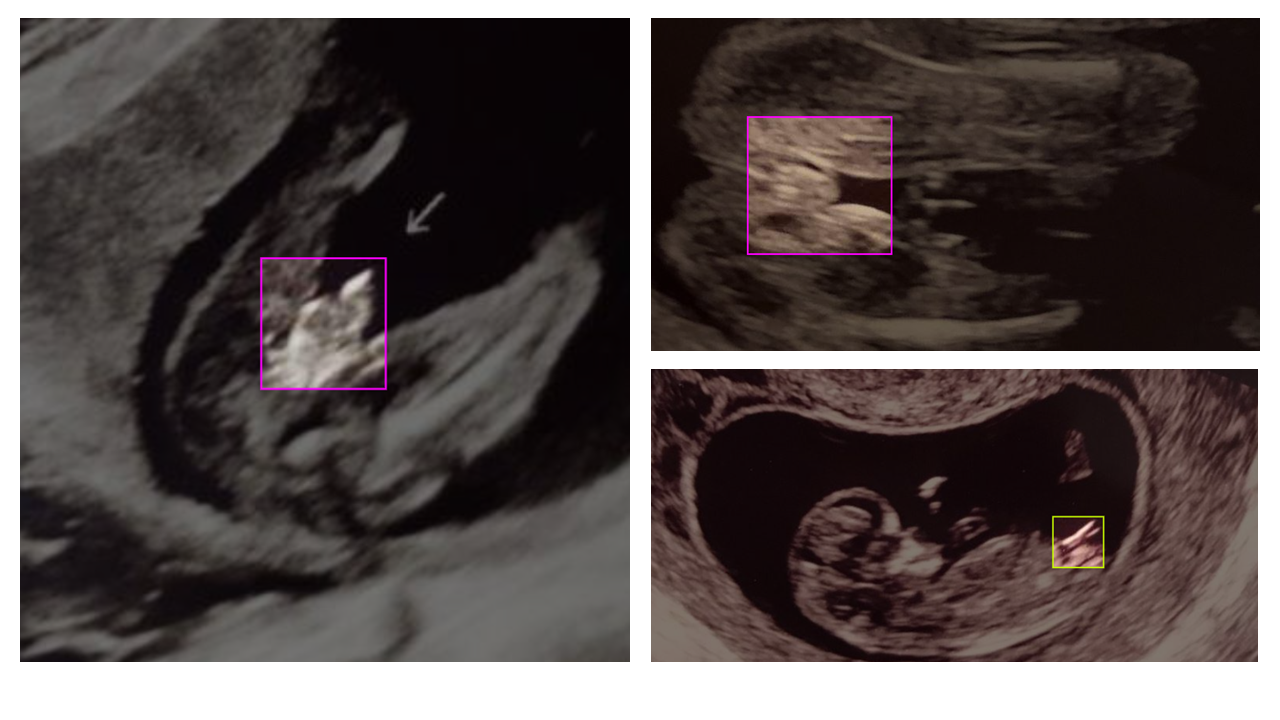}}
\caption{Sample Annotations}
\label{fig1}
\end{figure}

Fig.~\ref{fig1} depicts a few sample annotations.

\section{Experiments}
\label{sec:exp}

Several object detection architectures have been considered in this study~\cite{FasterRT, RetinaNet, EfficientDet, YOLOv5}. Table~\ref{table:archs} provides the details of various configurations of selected architectures. 

\begin{table}[!b]
\centering
\caption{Configuration of Object Detection Architectures\strut}
\label{table:archs}
\begin{tabular*}{0.6\textwidth}{ccc}
\toprule
Model & Backbone & \# params \\
\midrule
Faster R-CNN & ResNet-50-FPN-1x &  41 M\\
Faster R-CNN & ResNet-101-FPN-1x & 60 M \\
RetinaNet & ResNet-50-FPN-1x & 36 M\\
RetinaNet & ResNet-101-FPN-1x & 55 M\\
EfficientDet-D2 (AdvProp) & EfficientNet-B2 & 8.1 M  \\
EfficientDet-D3 (AdvProp) & EfficientNet-B3 & 12 M  \\
EfficientDet-D4 (AdvProp) & EfficientNet-B4 & 21 M  \\
EfficientDet-D5 (AdvProp) & EfficientNet-B5 & 34 M\\
YOLOv5L & YOLOv5L CSPDarknet & 46.6 M\\ 
\bottomrule
\end{tabular*}
\end{table}

\subsection{Training}

Models were trained using transfer learning from weights trained on the COCO dataset \cite{coco} for YOLOv5L and the ImageNet dataset \cite{deng2009imagenet} for all other models.  The models' hyperparameters were optimized by observing the validation set performance. All non-YOLOv5L models were trained for 20 epochs with pretrained frozen backbones and another 20 epochs with the entire parameters unfrozen, using the Adam optimizer~\cite{Adam}. Training for additional epochs did not provide an improvement. The SGD optimizer was used for the YOLOv5L model, and training was performed for 125 epochs with pretrained and unfrozen backbone parameters. All models were trained using a one-cycle learning rate scheduler~\cite{one:cycle}. The maximum learning rate was set to 0.01 for YOLOv5L, 0.003 for EfficientDet variants, and 0.0001 for other models. Layer discriminative learning rates were used for fine-tuning the pretrained backbones. For the EfficientDet variants, a range of $[3\times10^{-5},3\times10^{-4}]$ was used, while a range of $[1\times10^{-6},1\times10^{-5}]$ was used for the Faster R-CNN and RetinaNet models.

The following data augmentations were used in training: ShiftScaleRotate, RGBShift, RandomBrightnessContrast, RandomFlip, RandomSizedCrop, and Blur. An adversarial training procedure (AdvProp)~\cite{AdvProp} was used to enhance the performance of EfficientDet models. Additionally, YOLOv5 uses Mosaic~\cite{YOLOv4} and Mixup~\cite{MixUp}.

\subsection{Results}

In this section, we evaluate the performance of the models on the test set. The input image size after augmentation was fixed to $384$ pixels square in order to directly compare the models. The \textit{detection confidence} and $IoU$ \textit{threshold} parameters were both set to $0.5$.  Models are compared on average precision ($AP$), label Precision ($P$), label Recall ($R$), and label F1-score ($F1$). $AP_{0.5}$ is the average precision of all classes with bounding box predictions over ground truth of $IoU > 0.5$. $AP_{[0.5:0.95]}$ is the average of the average precision across all classes at $IoU$ between $0.5$ and $0.95$ with a step size of $0.05$. Table~\ref{table:test.res} shows the metrics on the test set.

\begin{table}[b]
\caption{Performance on the test set at $conf_{thr}=0.5$, $IoU_{thr}=0.5$}
\label{table:test.res}
\setlength\tabcolsep{0pt}
\begin{tabular*}{\columnwidth}{@{\extracolsep{\fill}}l ccccc}
\toprule
Model & $AP_{[0.5:0.95]}$ & $AP_{0.5}$ & $P$ & $R$ & $F1$ \\
\midrule
Faster R-CNN (ResNet 50) & 0.359 & 0.795 & 0.756 & 0.821 & 0.787\\
Faster R-CNN (ResNet 101) & 0.356 & 0.809 & 0.84 & 0.834 & 0.837\\
RetinaNet (ResNet 50) & 0.307 & 0.704 & 0.931 & 0.711 & 0.806\\
RetinaNet (ResNet 101) & 0.366 & 0.809 & 0.906 & 0.834 & 0.869\\
EfficientDet-D2 (AdvProp) & 0.371 & 0.796 & 0.961 & 0.821 & 0.886\\
EfficientDet-D3 (AdvProp) & 0.404 & 0.845 & 0.963 & 0.861 & 0.909\\
EfficientDet-D4 (AdvProp) & 0.387 & 0.808 & 0.933 & 0.834 & 0.881\\
EfficientDet-D5 (AdvProp) & 0.419 & 0.837 & 0.942 & 0.861 & 0.900\\
YOLOv5L & 0.448 & 0.853 & 0.962 & 0.855 & 0.905\\  
\bottomrule
\end{tabular*}
\end{table}

\subsection{Discussion}

\begin{figure}[!t]
\centerline{\includegraphics[scale=0.4]{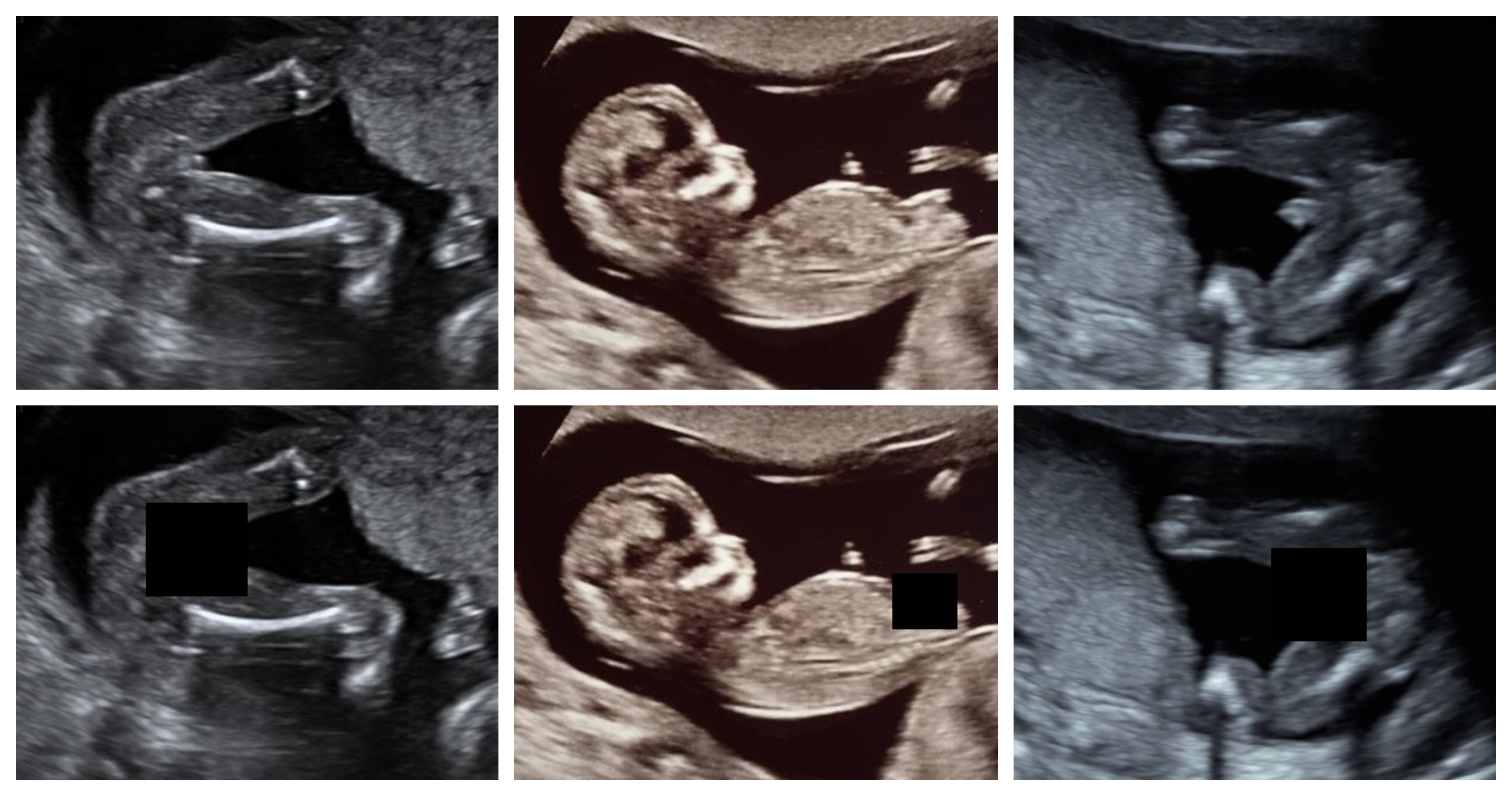}}
\caption{Sample detection and masking from the test set (YOLOv5L)}
\label{fig2}
\end{figure}

The results of the YOLOv5L and EfficientDet-D3 (AdvProp) models show a very high performance of over 90\% label F1-score, with the YOLOv5L model's higher $AP_{[0.5:0.95]}$ demonstrating a better average precision over different IoU thresholds. Fig.~\ref{fig2} depicts sample detection and masking on images from the test set using YOLOv5L. In order to further analyze the performance of the best performing YOLOv5L model, further investigation into False Positives and False Negatives will be made in the following sections.

\section{Minimizing False Negatives \& False Positives}
\label{sec:opt}

For any deep learning model, domain-shift between the training data and the real world is a necessary practical consideration. In object detection, this can potentially lead to problems caused by false negative and false positive detections.

For the proposed application, false negatives (FNs) are potentially extremely costly. If the model fails to mask the genital region, the gender of the fetus may be revealed. In addition, false positives (FPs) are potentially very harmful in practice. FPs correspond to incorrect detection and masking of areas that do not contain the regions of interest. This behavior is not desired as it may prevent the practitioner from performing a smooth scan by incorrectly masking areas in other fetal anatomical planes like the abdomen or the brain. 

\begin{figure}[!t]
\centerline{\includegraphics[scale=0.3]{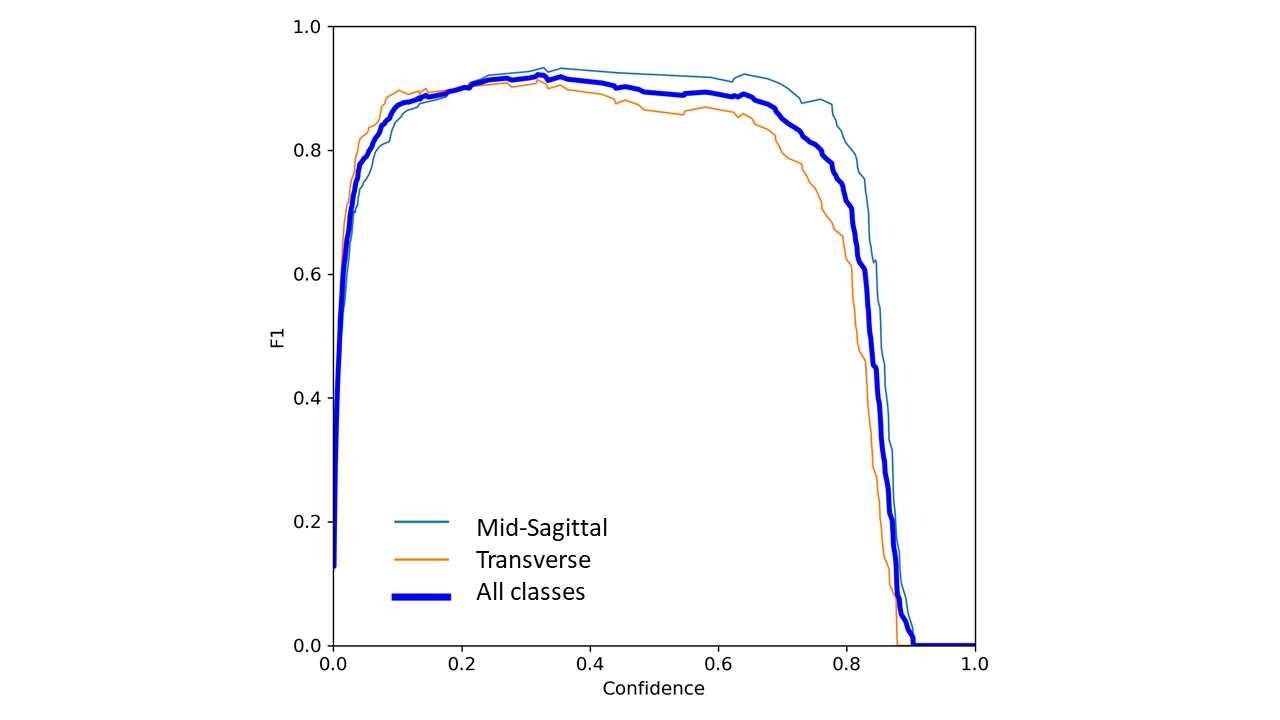}}
\caption{F1 detection confidence threshold curve for YOLOv5L}
\label{fig3}
\end{figure}

In order to enhance the performance of our best-performing model and particularly minimize FNs and FPs, the detection confidence and IoU thresholds have been investigated. The highest F1-score of 0.92 with $AP_{[0.5:0.95]}$ 0.458 and FPPI 0.006, defined as the rate of False Positive detections per frame \cite{5206631}, is found at detection confidence threshold 0.318 and IoU threshold 0.6~\cite{YOLOv5}, as shown in Fig.~\ref{fig3}.

\section{Masking Performance: Specialist Sonographer}
\label{sec:Sono}

In order to confirm the practical utility of our proposed system, a registered expert medical practitioner has provided guidance in the development of the detection masking functionality and further feedback on the final performance.

One hundred ROI-containing images from the test set were provided to the medical expert, first with ROIs masked by our system, and then subsequently, the same original (unmasked) images were provided. In both cases, the medical expert's task was to attempt to identify the gender of the fetus. In the case of the masked images, the gender identification rate was 5\%. The masking by our system was shown to provide a reduction in recall rates of 88\% for the male fetal gender identification and 97\% for the female gender identification.  However, it should be noted that all of the correct gender identifications were made on images with false negatives and therefore no masking was present. As such, the fetal gender within 100\% of the correctly masked images was not identifiable. This further impresses the requirement of reducing FN detections to negligible amounts.

It should be noted the data set comprises publicly available images and, as such, the results reported here are for demonstration purposes only and should not be considered indicative of performance in deployment. Further development and evaluation of the system under clinical conditions are therefore required.

\section{Further Work}
\label{sec:further}

\subsection{Post-processing to Reduce False Negatives in Real-Time Masking}

No substantial gaps of consecutive false negatives were observed in the proposed system when testing on multiple videos of fetal US scans. However, frames with false negatives were occasionally observed. In addition to the optimization measures stated in Section~\ref{sec:opt}, empirical testing has shown that the following two proposed approaches effectively reduce this problem; \textit{BBoxHold} and \textit{BBoxHoldSim}.

The \textit{BBoxHold} approach is quite simple: Whenever the model predicts a bounding box, the recently predicted bounding box is held for a few frames if no predictions are made. This approach has been found to reduce the potential flickering effect of occasionally missing predictions on true-positive frames. However, this approach may fail when the model does not predict anything for a sufficiently long enough time to reveal gender, depending on the hold time. \textit{BBoxHoldSim} approach tackles this problem by comparing the current frame with the most recent frame containing a bounding box prediction using Structural Similarity Index Measure (SSIM)~\cite{1284395}. If the similarity is greater than a threshold, \textit{BBoxHold} is activated.

The two approaches were tested on four sections of US videos comprising gender identifying frames, where the \textit{BBoxHold} and \textit{BBoxHoldSim} approaches were found to reduce FN rates by 56\% and 85\%, respectively, compared to the YOLOv5L model alone at the optimum F1 score detection confidence threshold of 0.318 and IoU threshold 0.6. However, further investigation into the combined detection threshold settings and these methods is required.

\subsection{US Video Test Dataset}

In order to aid further development of this system, a test dataset formed from consecutive video frames of several US fetal scan videos including gender determination frames is likely to be required. By doing so, the training can be tuned to maximally reduce any remaining false negatives and false positives in video streams, that are apparent when training on single non-consecutive gender-region frames.

\section{Conclusion}
\label{sec:con}

In this paper, a system for masking fetal gender in ultrasound scans during the early stages of pregnancy using deep learning-based object detection has been proposed. An F1-score of 92\% for detection and masking has been achieved on our dataset containing both true-positive and negative samples simulating a full fetal scan. In addition, further delay and Structural Similarity Index Measure methods have been investigated in order to reduce the levels of potential false negative detections that could otherwise reveal the fetal gender. We hope the integration of this approach into ultrasound systems will allow a more ubiquitous usage of ultrasound scans in regions currently affected by sex-selective abortions caused by fetal gender determination.

\bibliographystyle{unsrt}
\bibliography{ref}

\end{document}